\documentclass[11pt]{article}
\usepackage[final]{acl}

\usepackage{times}
\usepackage[T1]{fontenc}
\usepackage[utf8]{inputenc}

\usepackage{amsmath,amssymb,amsfonts}
\usepackage{latexsym}
\usepackage{pifont}

\usepackage{graphicx}
\usepackage{multirow}

\usepackage{booktabs,threeparttable,xcolor,colortbl}
\usepackage{tcolorbox}
\usepackage{algorithm}
\usepackage{algpseudocode}
\tcbuselibrary{listings, breakable}

\definecolor{headerblue}{HTML}{B7C5E6}
\definecolor{datacellgreen}{HTML}{D9EAD3}
\definecolor{datacellpink}{HTML}{FFD1DC}

\usepackage{threeparttable}
\usepackage[table]{xcolor}
\usepackage{booktabs}
\newcommand{\cmark}{\textcolor{green!60!black}{\ding{51}}}
\newcommand{\xmark}{\textcolor{red!70!black}{\ding{55}}}
\newcommand{\pmark}{\textcolor{orange!80!black}{\raisebox{0.2ex}{\texttildelow}}}

\definecolor{HeaderBlue}{RGB}{220,235,250}
\definecolor{RowAlt}{RGB}{245,248,255}
\definecolor{OurRow}{RGB}{235,250,235}

\definecolor{CaseGreenFrame}{RGB}{0,150,0}
\definecolor{CaseGreenBack}{RGB}{242,252,242}
\definecolor{CaseRedFrame}{RGB}{200,0,0}
\definecolor{CaseRedBack}{RGB}{255,242,242}

\newtcolorbox{casepanel}[3][]{%
  enhanced,
  breakable,
  sharp corners,
  boxrule=0.9pt,
  colback=#2,
  colframe=#3,
  left=10pt,right=10pt,top=10pt,bottom=10pt,
  colbacktitle=#3,
  coltitle=black,
  fonttitle=\bfseries,
  title=#1,
}

\usepackage[T1]{fontenc}
\usepackage{booktabs} 
\usepackage{tcolorbox}
\tcbuselibrary{listings,skins,breakable}
\usepackage{pifont} 
\usepackage{amssymb}
\usepackage{xcolor}

\usepackage[utf8]{inputenc}

\usepackage{microtype}

\usepackage{inconsolata}

\usepackage{graphicx}
\usepackage[table]{xcolor}  % <-- load xcolor ONCE, with [table]
\usepackage{colortbl}
\tcbuselibrary{listings,skins,breakable}
\title{Tree-of-Evidence: Efficient "System 2" Search for Faithful Multimodal Grounding}

\author{\textbf{Micky C. Nnamdi$^{\spadesuit}$, Benoit L. Marteau$^{\spadesuit}$, Yishan Zhong$^{\spadesuit}$, J. Ben Tamo$^{\spadesuit}$,} \\ and \textbf{May D. Wang$^{\spadesuit}$}\\[0.5em]
$^{\spadesuit}$Georgia Institute of Technology} 

\begin{document}
\maketitle
\begin{abstract}
Large Multimodal Models (LMMs) achieve state-of-the-art performance in high-stakes domains like healthcare, yet their reasoning remains opaque. Current interpretability methods, such as attention mechanisms or post-hoc saliency, often fail to faithfully represent the model's decision-making process, particularly when integrating heterogeneous modalities like time-series and text. We introduce Tree-of-Evidence (ToE), an inference-time search algorithm that frames interpretability as a discrete optimization problem. Rather than relying on soft attention weights, ToE employs lightweight Evidence Bottlenecks that score coarse groups or units of data (e.g., vital-sign windows, report chunks) and performs a beam search to identify the compact evidence set required to reproduce the model’s prediction. We evaluate ToE across six tasks spanning three datasets and two domains: four clinical prediction tasks on MIMIC-IV, cross-center validation on eICU, and non-clinical fault detection on LEMMA-RCA. ToE produces auditable evidence traces while maintaining predictive performance, retaining over 98\% of full-model AUROC with as few as five evidence units across all settings. Under sparse evidence budgets, ToE achieves higher decision agreement and lower probability fidelity error than other approaches. Qualitative analyses show that ToE adapts its search strategy: it often resolves straightforward cases using only vitals, while selectively incorporating text when physiological signals are ambiguous. ToE therefore provides a practical mechanism for auditing multimodal models by revealing which discrete evidence units support each prediction. 
\end{abstract}

\section{Introduction}
\label{sec:intro}
Multimodal predictors, such as Large Multimodal Models (LMMs), have achieved remarkable performance by fusing heterogeneous data streams, including text, time series, and imaging, into unified representations~\cite{chen2024evolution, huang2024large, tu2024towards}. However, as these models grow in complexity, their decision-making processes become increasingly opaque~\cite{rudin2019stop, wornow2023shaky}. In high-stakes domains like healthcare, "black box" accuracy is insufficient; deployment requires auditable reasoning where a model’s prediction explicitly traces back to specific, verifiable pieces of evidence~\cite{rudin2019stop}. 

\begin{figure*}
    \centering
    \includegraphics[width=\textwidth, height=230px]{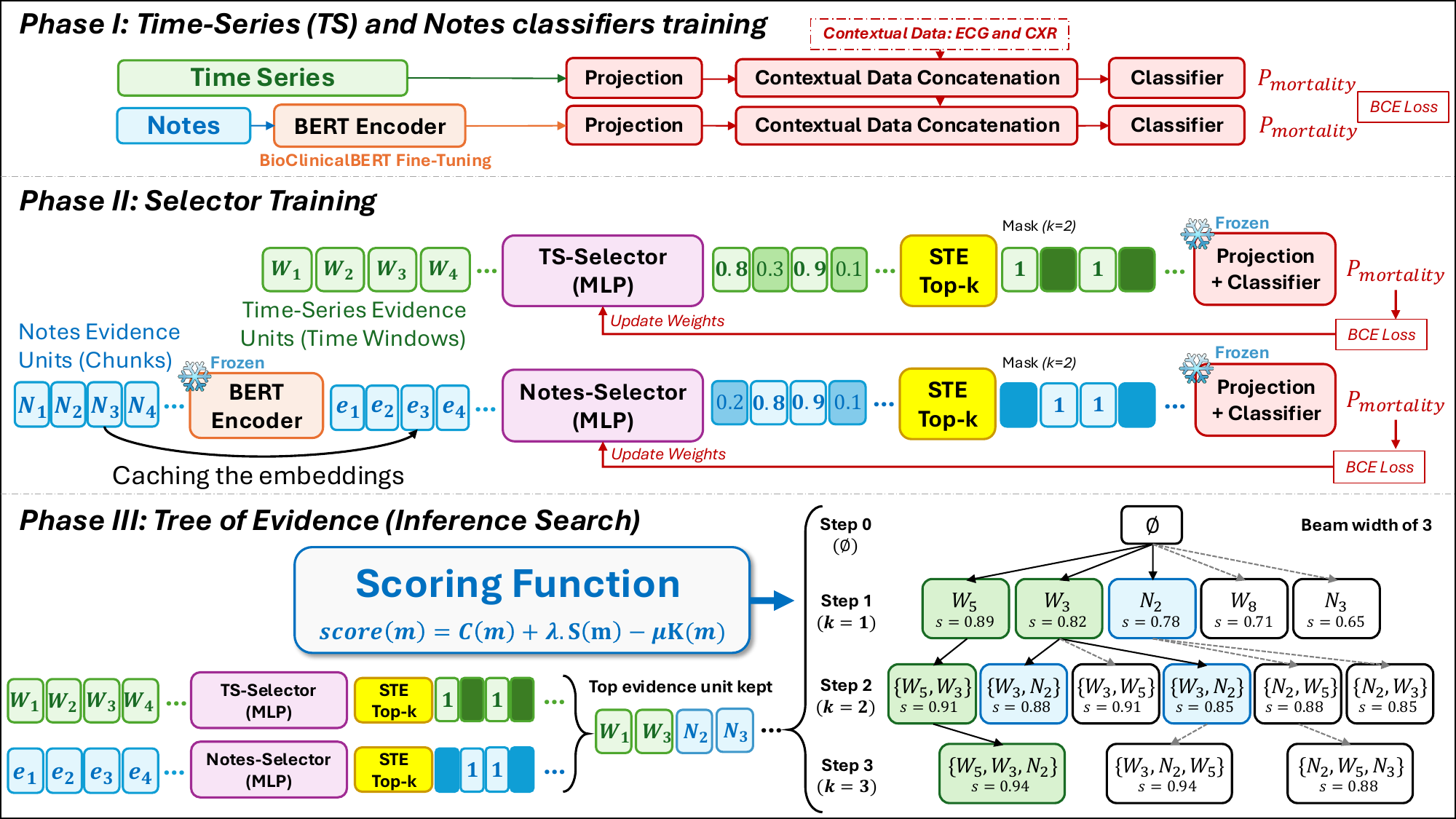}
    \caption{\textbf{Overview of the Tree-of-Evidence (ToE)} Framework. \textbf{Phase I}: Modality-specific classifiers are trained independently, with BioClinicalBERT~\cite{alsentzer2019publicly} encoding notes and contextual data (CXR/ECG) concatenated as fixed priors. \textbf{Phase II}: Lightweight MLP selectors learn to score evidence units using Straight-Through Estimator (STE) top-k masking with frozen encoders. \textbf{Phase III}: At inference, beam search iteratively constructs compact evidence set by optimizing the scoring function, balancing decision agreement, probability stability, and sparsity.}
    \label{fig:overview}
\end{figure*}

Current interpretability methods often fail to meet this standard. Attention-based heatmaps are frequently unfaithful to the actual logic of the model~\cite{wiegreffe-pinter-2019-attention, jain2019attention}, while post-hoc explanation methods provide approximations rather than guarantees~\cite{rudin2019stop}. Concept Bottleneck Models (CBMs) offer a step forward by aligning hidden states with human-interpretable concepts~\cite{koh2020concept, vandenhirtz2024stochastic}. Yet, CBMs typically require predefined concept annotations and remain static during inference, failing to adaptively search for evidence when data is ambiguous or synergistic. Rationale extraction methods aim to solve this by selecting a subset of input features that are sufficient for the prediction~\cite{lei-etal-2016-rationalizing, deyoung2020eraser}. Yet, existing rationale methods are typically limited to single modalities, mainly text, and rely on greedy selection strategies that fail to capture the synergistic dependencies between different data types~\cite{xu2024comprehensive}. For instance, a radiology note of pulmonary edema can reframe a drop in $SpO_{2}$ as volume overload — a cross-modal link that unimodal methods miss.

To bridge this gap, we introduce \textbf{Tree-of-Evidence (ToE)}, an inference-time search algorithm for multimodal grounding. Inspired by deliberative branching procedures like Tree-of-Thoughts~\cite{yao2023tree}, ToE treats interpretability as a \emph{discrete search problem} over meaningful evidence units. We use "System~2" to denote this multi-step, deliberative search process, in which the algorithm explicitly evaluates and scores candidate evidence combinations via beam search, in contrast to "System~1" single-pass heuristics such as greedy top-$k$ ranking by individual unit scores. Crucially, we structure the multimodal space into two distinct roles: (1) Global Context (e.g., baseline pathology from Chest X-Ray (CXR)/ Electrocardiogram (ECG)), which serves as a fixed prior, and (2) Searchable Evidence (e.g., dynamic Vitals and Notes), which is actively selected. Instead of relying on soft attention weights, we first train lightweight Evidence Bottlenecks (EB)  that score coarse units of data--hourly windows of Intensive Care Unit (ICU) time-series and radiology report chunks. At inference time, ToE performs a beam search to construct a compact evidence set that preserves the full-input decision, explicitly trading off (i) agreement with the original prediction, (ii) stability of the predicted probability, and (iii) evidence sparsity. This separation allows the search to focus on "what changed" (dynamic evidence) while remaining grounded in "who the patient is" (global context). The result is an auditable trace of how evidence is accumulated to justify a decision. 

We evaluate ToE across six tasks spanning three datasets and two domains: four clinical prediction tasks on MIMIC-IV~\cite{johnson2024mimic, johnson2023mimic, goldberger2000physiobank, gow2023mimic}, cross-center validation on eICU (208~hospitals)~\cite{pollard2018eicu}, and non-clinical fault detection on LEMMA-RCA~\cite{zheng2024lemma}. Our experiments demonstrate that ToE yields discrete rationales that (i) remain sufficient for the model's decision under strict evidence budgets, (ii) exhibit strong decision agreement with the full-input prediction, and (iii) provide an auditable trace of the search process that can be inspected by domain experts. Our contribution can be summarized as:

\begin{enumerate}
    \item \textbf{Model-faithful multimodal grounding via inference-time search.} We formulate grounding as selecting a compact multimodal evidence set that reproduces the full-input model’s decision and confidence, and we propose Tree-of-Evidence (ToE) to solve this with an auditable search trace.
    
    \item \textbf{Bottleneck-guided discrete evidence units.} We develop lightweight \emph{Evidence Bottlenecks} that score clinically meaningful, coarse-grained units (hourly windows; report chunks) and provide efficient heuristics for search, while incorporating CXR/ECG signals as \emph{context-only} features rather than searchable evidence.
    % \item \textbf{Faithfulness-first evaluation under evidence budgets.}
    % We evaluate explanations using sufficiency, comprehensiveness, and probability-agreement metrics under strict evidence constraints, and provide ablations and analyses showing ToE better preserves full-input behavior at a given sparsity than greedy or attribution-based baselines.

    \item \textbf{Comprehensive faithfulness evaluation under evidence budgets.} We evaluate explanations using sufficiency, comprehensiveness, and probability-agreement metrics under strict evidence constraints across three datasets, six tasks, and two domains. We compare against Local Interpretable Model-Agnostic Explanations (LIME), SHapley Additive exPlanations (SHAP), Concept Bottleneck Models, gradient saliency, and LLMs up to 70B parameters, and provide ablations showing ToE better preserves full-input behavior at a given sparsity than all baselines.
\end{enumerate}

\begin{table}[t]
\
\centering
\small
\setlength{\tabcolsep}{6pt}
\caption{\textbf{Comparison of Interpretability Frameworks.} 
ToE is distinct in offering an \textit{auditable trace} (search history) over \textit{multimodal} hard evidence.}
\label{tab:comparison}

\rowcolors{2}{RowAlt}{white}
\resizebox{\linewidth}{!}{
\begin{tabular}{lcccccc}
\toprule
\rowcolor{HeaderBlue}
\textbf{Method} 
& \textbf{Type} 
& \textbf{Hard Evidence?} 
& \textbf{Multimodal?} 
& \textbf{Faithfulness?} 
& \textbf{Auditable Trace?} \\
\midrule

Attention Weights & Intrinsic (Soft) & \xmark & \pmark & \xmark & \xmark \\
Gradient Saliency & Post-hoc & \xmark & \pmark & \xmark & \xmark \\
LIME / SHAP & Post-hoc Surrogate & \xmark & \pmark & \pmark & \xmark \\
Concept Bottleneck & Intrinsic (Concepts) & \cmark & \pmark & \pmark & \pmark \\

\midrule
\rowcolor{OurRow}
\textbf{Tree-of-Evidence (Ours)} 
& \textbf{Inference Search} 
& \textbf{\cmark} 
& \textbf{\cmark} 
& \textbf{\cmark} 
& \textbf{\cmark} \\
\bottomrule
\end{tabular}}
\end{table}

\section{Related Work}

\textbf{Faithful Rationale Extraction.}
% Rationale extraction seeks a \emph{subset} of input features that justifies a prediction~\cite{lei-etal-2016-rationalizing}. A central evaluation goal is \emph{faithfulness}: the selected rationale should be causally tied to the model’s behavior rather than merely plausible to humans~\cite{jacovi2020towards}. Common operationalizations include \emph{Sufficiency} (does the model make the \emph{same} prediction when restricted to the rationale?) and \emph{Comprehensiveness} (does removing the rationale change the prediction?)~\cite{deyoung2020eraser}. Information Bottleneck-style methods encourage concise rationales by penalizing information passed from the input~\cite{paranjape2020information}, but are most often studied in unimodal settings (typically text). Recent work further emphasizes that faithfulness metrics can be sensitive to evaluation design and may yield misleading comparisons if applied naively~\cite{chan-etal-2022-comparative, edin-etal-2025-normalized}.
% Our work adopts these faithfulness principles in the \emph{teacher-faithful} sense: we seek evidence sets that preserve the \emph{full-input model’s decision and confidence}, and we extend this evaluation to multimodal clinical inputs by defining discrete evidence units over structured time-series windows and radiology-report chunks under strict evidence budgets.

Rationale extraction seeks a subset of input features that justifies a prediction~\cite{lei-etal-2016-rationalizing}. A central evaluation goal is \emph{faithfulness}: the selected rationale should be causally tied to the model's behavior rather than merely plausible to humans~\cite{jacovi2020towards}. Common operationalizations include \emph{Sufficiency} (does the model make the same prediction when restricted to the rationale?) and \emph{Comprehensiveness} (does removing the rationale change the prediction?)~\cite{deyoung2020eraser}. Post-hoc attribution methods such as LIME and SHAP approximate feature importance via local surrogates or Shapley values, but provide no hard selection mechanism and can yield unstable explanations. Information Bottleneck-style methods encourage concise rationales by penalizing information passed from the input~\cite{paranjape2020information}, but are most often studied in unimodal settings. Concept Bottleneck Models (CBMs)~\cite{koh2020concept, vandenhirtz2024stochastic} align hidden states with human-interpretable concepts, yet require predefined annotations and remain static at inference time. Recent work further emphasizes that faithfulness metrics can be sensitive to evaluation design~\cite{chan-etal-2022-comparative, edin-etal-2025-normalized}. Table~\ref{tab:comparison} summarizes these distinctions: ToE is the only framework that combines hard evidence selection, multimodal support, faithfulness guarantees, and an auditable search trace.

\textbf{Search-Based Reasoning and Interpretability.}
Systematic search procedures such as Tree-of-Thoughts~\cite{yao2023tree} apply branching strategies to improve reasoning, typically in the token-generation space. More closely related are methods that apply search in the \emph{evidence selection} space. In computer vision tasks, Shitole et al.~(2021) use beam search to identify diverse attention maps that are individually sufficient for classification~\cite{shitole2021one}. Zhou and Shah show that standard faithfulness objectives (e.g., sufficiency/comprehensiveness) can be directly optimized and propose search-based explainers~\cite{zhou2023solvability}, raising the question of what should distinguish new methods beyond metric optimization.
\textbf{ToE} is motivated by these insights but differs in goal and setting: we search for a \emph{compact, decision- and probability-preserving} subset of \emph{multimodal clinical evidence units}, guided by learned Evidence Bottlenecks that efficiently propose and score candidate units. This contrasts with model-agnostic approaches such as Anchors~\cite{ribeiro2018anchors} and counterfactual explanations~\cite{wachter2017counterfactual}, which typically require extensive perturbation/sampling or additional optimization at query time rather than using jointly trained unit-level selectors.

\textbf{Multimodal Learning and Explainability in Healthcare.}
Integrating heterogeneous clinical data remains a core challenge in medical AI~\cite{acosta2022multimodal}. Many multimodal architectures combine text encoders (e.g., BERT-style models) with structured time-series encoders (e.g., LSTMs or Transformers) via late fusion, gating, or cross-attention~\cite{huang2020fusion, seki2021machine, golas2018machine}. While these designs can improve predictive performance, explanations are often provided via modality-specific post-hoc attributions (e.g., token saliency or feature importance) and rarely yield a \emph{single cross-modal evidence set} that can be audited end-to-end.
In contrast, our approach introduces a discrete evidence-selection layer over multimodal representations: ToE constructs an auditable evidence trace over \emph{time-series windows} and \emph{radiology report chunks}, enabling teacher-faithfulness checks via sufficiency, comprehensiveness, and probability-agreement metrics without constraining the underlying predictive backbone.

\section{Method}
\label{sec:method}
We formulate interpretable clinical prediction as a \emph{search problem}. Our framework, \textbf{ToE}, separates the reasoning process into two stages:
(1) learn efficient, differentiable heuristics for evidence scoring via \emph{EB}, and
(2) perform an inference-time discrete search to identify a compact, high-scoring evidence set required for a robust diagnosis. We represent an overview of our approach in Figure~\ref{fig:overview}.

\subsection{Problem Setup and Evidence Units}
We define a binary classification task $y\in\{0,1\}$ over an observation window $[t_0, t_0+\Delta]$ (default $\Delta{=}24$h). The input space $\mathcal{X}$ consists of two \emph{searchable} modalities and two \emph{context} modalities. While we present the formulation using ICU data as a running example, the framework applies to any setting where inputs can be decomposed into discrete units across one or more modalities; we evaluate on non-ICU settings in Section~\ref{sec:results}.

\paragraph{Structured ICU Time Series (searchable evidence).}
We represent ICU measurements (vital signs and lab values) as a fixed-length sequence $\mathbf{x}^{\text{ts}} = (x^{\text{ts}}_1, \ldots, x^{\text{ts}}_T)$ with $T = 24$ hourly bins and $x^{\text{ts}}_t \in \mathbb{R}^D$. Each bin contains summary statistics (e.g., mean, min, max) over vitals and labs in that hour, along with missingness indicators. Evidence Units are the discrete time windows $\{W_t\}_{t=1}^{T}$ corresponding to these bins.

\paragraph{Radiology Reports (searchable evidence).}
Let $\mathbf{x}^{\text{note}}$ be the concatenation of all radiology report text within the window $[t_0, t_0 + \Delta]$. We segment this text into a sequence of chunks $(c_1, \ldots, c_M)$ (e.g., 3-sentence segments), padded or truncated to a fixed length $M_{\max}$. Let $\mathbf{a} \in \{0, 1\}^{M_{\max}}$ be a presence mask indicating valid (non-padding) chunks. Evidence Units are the discrete chunks $\{N_j\}_{j=1}^{M_{\max}}$.

\paragraph{CXR and ECG Context (Global Priors).}
To ground the search in the patient's broader physiological state, we include fixed context vectors that are not subject to selection: (i)~$\mathbf{x}^{\text{cxr}} \in \mathbb{R}^{D_{\text{cxr}}}$, a label vector from the most recent CXR (e.g., CheXpert) with indicator $\texttt{has\_cxr}$; and (ii)~$\mathbf{x}^{\text{ecg}} \in \mathbb{R}^{D_{\text{ecg}}}$, machine measurements from the most recent ECG with indicator $\texttt{has\_ecg}$.

We deliberately model these signals as fixed priors rather than searchable units to mirror clinical reasoning. CXR and ECG typically represent the patient's baseline physiological state (chronic/background), whereas notes and vitals represent acute evolution (dynamic). By conditioning the search on fixed context, ToE forces the model to identify dynamic evidence that explains the outcome given the patient's baseline risk, preventing the search from wasting budget on static confirmational signals.

\paragraph{Evidence Set.}
We formally define an explanation as a tuple of indices $E = (E^{\text{ts}}, E^{\text{note}})$, where $E^{\text{ts}} \subseteq \{1, \ldots, T\}$ indexes selected time windows and $E^{\text{note}} \subseteq \{1, \ldots, M_{\max}\}$ indexes selected note chunks.

\subsection{Evidence Bottleneck Predictors}
\label{sec:eb}
We employ a modular architecture with two EB streams, corresponding to the searchable modalities ($\mathbf{x}^{\text{ts}}$ and $\mathbf{x}^{\text{note}}$). Each stream consists of: (i)~a \emph{Selector} that scores discrete evidence units to produce a hard top-$k$ mask; and (ii)~a \emph{Predictor} that estimates the diagnosis using only the selected subset. This separation ensures that the model cannot ``cheat'' by accessing information it has not explicitly selected.

\subsubsection{Differentiable Top-$k$ Selector}
\label{sec:selector}
Let $U = \{u_1, \ldots, u_n\}$ be the set of evidence units (time windows or chunk embeddings). A lightweight MLP selector $f_\theta$ assigns a scalar relevance score $s_i = f_\theta(u_i)$ to each unit. For variable-length inputs (notes), we enforce validity by setting $s_i = -\infty$ wherever the presence mask $a_i = 0$, ensuring padding is never selected.
 
To enable end-to-end training with discrete selection, we utilize the Straight-Through Estimator (STE). We compute a hard top-$k$ mask $\mathbf{m} = \text{TopK}(\mathbf{s}, k) \in \{0, 1\}^n$ for the forward pass, but approximate gradients via a softmax relaxation $\tilde{\mathbf{m}} = \text{softmax}(\mathbf{s}/ \tau_{\mathrm{STE}})$ during backpropagation:

\begin{equation}
    \hat{\mathbf{m}} = \mathbf{m} - \text{sg}(\tilde{\mathbf{m}}) + \tilde{\mathbf{m}},
    \label{eq:ste}
\end{equation}

where $\tau_{\mathrm{STE}}$ is the STE temperature (default $\tau_{\mathrm{STE}}$ = 1.0; see Appendix \ref{app:ste_temperature}), $\text{sg}(\cdot)$ denotes the stop-gradient operator. This allows the selector to update its ranking logic $\theta$ based on the downstream predictor's performance.

The STE introduces a forward-backward gradient mismatch by construction. Our two-phase training design mitigates this: in Phase~I, the predictor trains with all evidence selected ($k = T$), so the STE is never invoked; in Phase~II, the predictor is frozen, and only the selector MLP (98K of 109M total parameters) is updated. Because the frozen predictor's weights are fixed, the selector needs only to learn a correct \emph{ranking} of units, which units the predictor finds most informative, rather than propagating calibrated classification gradients end-to-end. Gradient mismatch affects magnitude but not ordering, preserving the ranking objective. Empirically, sufficiency Area Under the Receiver Operating Characteristic curve (AUROC) varies less than 1\% across a 50$\times$ temperature range ($\tau_{\mathrm{STE}} \in \{0.1, 5.0\}$; Appendix~\ref{app:ste_temperature}).

\subsubsection{Modality-Specific Encoders}
\textbf{Time-Series Stream.} The selector scores raw feature vectors $u_t = x^{\text{ts}}_t$. We apply the mask element-wise, $\tilde{x}^{\text{ts}}_t = \hat{m}^{\text{ts}}_t \cdot x^{\text{ts}}_t$, effectively zeroing out non-selected hours. The sequence is encoded via a Bidirectional Gated Recurrent Unit (GRU) to obtain a final representation $\mathbf{v}^{\text{ts}}$ (concatenated hidden states). While we employ a GRU for computational efficiency, our framework is model-agnostic and compatible with continuous-time encoders such as Latent ODEs~\cite{rubanova2019latent}. Finally, we inject global context by concatenating the projected context vectors:
\begin{equation}
    \mathbf{z}^{\text{ts}} = [\mathbf{v}^{\text{ts}};\; \psi^{\text{cxr}}(\mathbf{x}^{\text{cxr}});\; \psi^{\text{ecg}}(\mathbf{x}^{\text{ecg}})],
    \label{eq:ts_context}
\end{equation}
where $\psi$ are lightweight projection MLPs. A classifier $g^{\text{ts}}_\phi$ maps $\mathbf{z}^{\text{ts}}$ to the logit $\ell^{\text{ts}}$.

\textbf{Notes Stream.}
We embed text chunks using a frozen BioClinicalBERT encoder, $e_j = \text{BERT}(c_j)_{[\text{CLS}]}$. The selector scores these embeddings to obtain a mask $\hat{\mathbf{m}}^{\text{note}}$. The predictor computes a masked mean pool over the selected valid chunks:

\begin{equation}
    \mathbf{v}^{\text{note}} = \frac{\sum_j \hat{m}^{\text{note}}_j \, a_j \, \phi^{\text{note}}(e_j)}{\sum_j \hat{m}^{\text{note}}_j \, a_j + \epsilon},
    \label{eq:notes_pool}
\end{equation}
where $\phi^{\text{note}}$ is a learnable projection MLP. Similar to the time-series stream, we inject global context:

\begin{equation}
    \mathbf{z}^{\text{note}} = [\mathbf{v}^{\text{note}};\; \psi^{\text{cxr}}(\mathbf{x}^{\text{cxr}});\; \psi^{\text{ecg}}(\mathbf{x}^{\text{ecg}})],
    \label{eq:notes_context}
\end{equation}
and pass $\mathbf{z}^{\text{note}}$ to a classifier $g^{\text{note}}$ to produce the logit $\ell^{\text{note}}$.

\subsubsection{Training and Inference Fusion}
\label{sec:fusion}
We train the streams separately using class-balanced Binary Cross-Entropy. This isolation ensures that each modality learns independent grounding logic without over-relying on the other.

At inference, we fuse the streams via logit summation. We define the predicted probability for binary evidence masks $\mathbf{m}^{\text{ts}}$ and $\mathbf{m}^{\text{note}}$ as:
\begin{equation}
    p(\mathbf{m}^{\text{ts}}, \mathbf{m}^{\text{note}}) = \sigma\!\left(\ell^{\text{ts}}(\mathbf{m}^{\text{ts}}) + \ell^{\text{note}}(\mathbf{m}^{\text{note}})\right),
    \label{eq:fusion}
\end{equation}

where $\ell(\cdot)$ denotes the logit output of a stream given a specific mask. We define the \emph{Full-Input Decision} as the prediction using all available units ($\mathbf{m} = \mathbf{1}$), denoted as $p_{\text{full}} = p(\mathbf{1}^{\text{ts}}, \mathbf{1}^{\text{note}})$ with predicted class $\hat{y}_{\text{full}}$.

\subsection{Faithfulness Evaluation}
\label{sec:faith}
We quantify interpretability using the Evaluating Rationales And Simple English Reasoning (ERASER) benchmark standards for faithfulness~\cite{deyoung2020eraser}.

\paragraph{Sufficiency.} Measures if the selected evidence is adequate to reproduce the prediction.
We report the model's performance (AUROC, Area Under the Precision-Recall Curve or AUPRC) when masking out all non-selected units (i.e., keeping only the top-$k$ evidence).

\paragraph{Comprehensiveness.} Measures if the model relies on the selected evidence. We calculate the drop in confidence for the \emph{originally predicted class} $\hat{y}_{\text{full}}$ when the selected evidence is removed.
Let $m_{\text{sel}}$ be the selected evidence mask and $m_{\text{rem}} = \mathbf{1} - m_{\text{sel}}$ be the complement. We compute:

\begin{equation}
    \Delta_{\text{comp}} = \frac{1}{N} \sum_{i=1}^{N} \left[ \Pr(\hat{y}^{(i)}_{\text{full}} \mid \mathbf{1}) - \Pr(\hat{y}^{(i)}_{\text{full}} \mid \mathbf{m}^{(i)}_{\text{rem}}) \right].
    \label{eq:comp}
\end{equation}
A higher $\Delta_{\text{comp}}$ indicates that the model's prediction relied heavily on the removed evidence.

\subsection{Tree-of-Evidence (ToE): Inference-Time Search}
\label{sec:toe}

Standard top-$k$ selection is brittle because it assumes evidence units are independent. However, clinical evidence is often synergistic (e.g., a radiology finding of alveolar edema contextualizes a subsequent drop in $SpO_{2}$). To address this, we propose ToE, a discrete beam search algorithm that identifies a compact evidence set to reproduce the full-input decision. Following the terminology introduced in Section~\ref{sec:intro}, this constitutes the ``System~2'' component of our framework: a multi-step deliberative search that explicitly evaluates candidate evidence combinations, in contrast to ``System~1'' single-pass greedy ranking.

\subsubsection{Search Space and Candidates}
A search state is a pair of binary masks $\mathbf{m} = (\mathbf{m}^{\text{ts}}, \mathbf{m}^{\text{note}})$. To keep the search tractable, we restrict actions to the top-$N$ candidates per modality (ranked by selector scores) to control computation.

\subsubsection{Search Objective}
We seek a state that maximizes confidence in the original decision while minimizing evidence cost. For a state $\mathbf{m}$, we define the scoring function:

\begin{align}
    C(\mathbf{m}) &= \Pr(\hat{y}_{\text{full}} \mid \mathbf{m}), \label{eq:agreement} \\
    S(\mathbf{m}) &= 1 - |p_{\text{full}} - p(\mathbf{m})|, \label{eq:stability} \\
    K(\mathbf{m}) &= \|\mathbf{m}^{\text{ts}}\|_0 + \|\mathbf{m}^{\text{note}}\|_0, \label{eq:sparsity} \\
    \text{score}(\mathbf{m}) &= C(\mathbf{m}) + \lambda\, S(\mathbf{m}) - \mu\, K(\mathbf{m}), \label{eq:score}
\end{align}
where $C$ is the model's confidence in the full-input predicted class (\emph{Sufficiency}), $S$ encourages probability stability (\emph{Calibration}), and $K$ penalizes evidence cost.

We define the stability term $S(\mathbf{m})$ in probability space (Eq.~\ref{eq:stability}) rather than logit space. This choice reflects three considerations. First, near $p = 0$ or $p = 1$, where most ICU patients fall, given class prevalences of 7--14\%, large logit deviations produce negligible probability changes; probability-space stability appropriately assigns low cost to these clinically irrelevant shifts. Second, the resulting metric is bounded in $[0, 1]$ and directly interpretable as ``mortality risk shifted by $X$ percentage points.'' Third, it is numerically stable, avoiding the divergences that logit-space distances exhibit near saturation. By including the stability term, the search does not merely maximize confidence (which could lead to selecting evidence that inflates a prediction) but explicitly aims to match the calibration of the full model. This ensures the selected evidence is not just ``sufficient'' in isolation, but faithful to the model's complete decision.

\subsubsection{Algorithm and Efficiency}
The search proceeds as follows (Algorithm~\ref{alg:toe}, Appendix~\ref{app:toe_algo}):
\begin{enumerate}
    \item \textbf{Initialization:} Start with an empty evidence set.
    \item \textbf{Expansion:} At each step, generate candidate states by adding exactly one unit from the candidate list $\mathcal{W} \cup \mathcal{N}$.
    \item \textbf{Pruning:} Evaluate candidates via frozen EB predictors and retain the top-$B$ states (Beam Width).
    \item \textbf{Termination:} Stop if a state meets sufficiency thresholds ($\tau_{\text{conf}}, \tau_{\text{suff}}$) or max steps are reached.
\end{enumerate}
Note that beam search finds high-scoring evidence sets under the scoring heuristic (Eq.~\ref{eq:score}), not globally optimal ones. Exhaustive enumeration confirms optimality gaps below 0.001 AUROC at small $k$ on MIMIC-IV (Appendix~\ref{app:optimality}).

\textbf{Efficiency via Caching:}
Since the BERT backbone is frozen, we cache the embeddings $\{e_j\}$ for all note chunks once per patient. During search, state evaluation requires only lightweight pooling and MLP passes, making ToE computationally efficient and suitable for deployment.

\section{Experiment}
\label{sec:results}
\begin{figure}
    \centering
    \includegraphics[width=\linewidth]{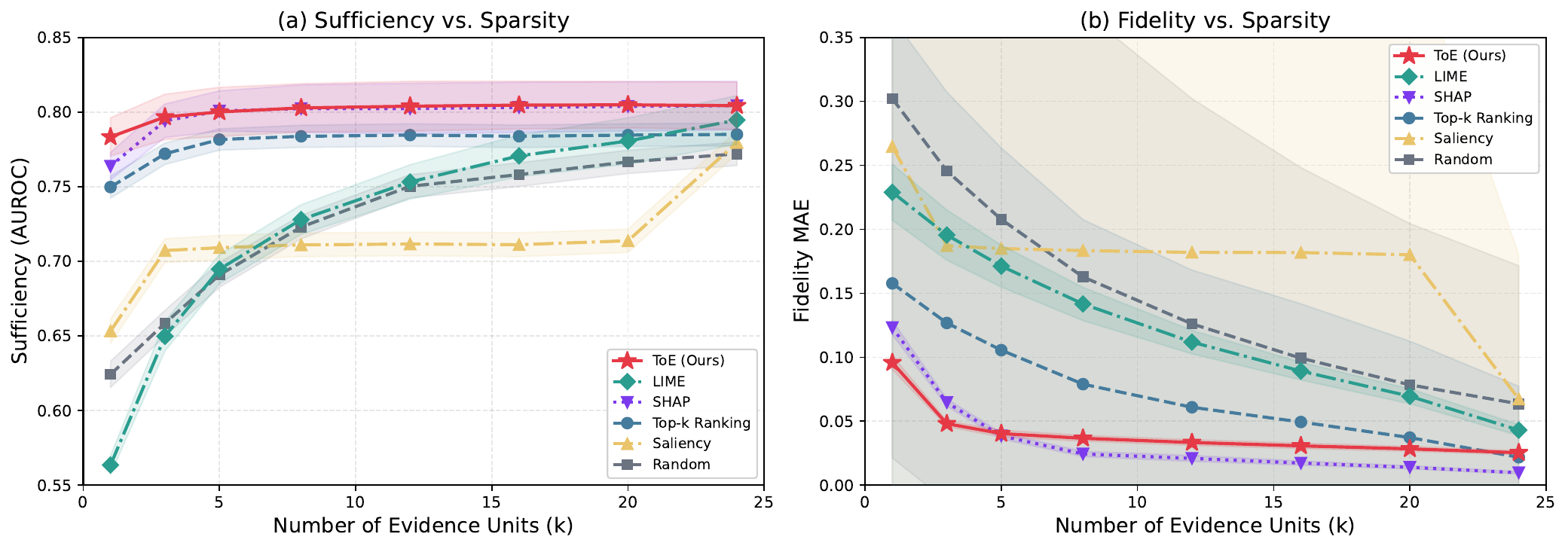}
    \caption{\textbf{The Faithfulness-Sparsity Frontier.} Performance across evidence budgets $k$ on MIMIC-IV (E1: In-Hospital Mortality, 5 seeds). \textbf{(a)~Sufficiency:} ToE (Red $\star$) matches the full model's predictive power ($\text{AUROC} \approx 0.80$) with as few as $k{=}5$ units. \textbf{(b)~Fidelity:} ToE achieves the lowest Fidelity MAE at sparse budgets ($k\leq5$), reducing error by ${>}50\%$ compared to Top-$k$ Ranking (Blue $\bullet$) and Saliency (Gold $\blacktriangle$) at sparse budgets, proving it captures the model’s actual confidence rather than just label correlations.}
    \label{fig:pareto_frontier}
\end{figure}

\subsection{Dataset and Implementation Details}
\paragraph{Dataset and Cohort.}
\paragraph{MIMIC-IV.}
Our primary evaluation uses the MIMIC-IV dataset~\cite{johnson2024mimic, johnson2023mimic}. The cohort consists of adult patients with at least 24~hours of ICU observation data. The final dataset comprises $N = 74{,}829$ unique ICU stays, split into training ($N = 52{,}597$), validation ($N = 11{,}053$), and testing ($N = 11{,}179$). We evaluate on four prediction tasks: (E1)~Hospital Mortality (prevalence 11.5\%), (E2)~Long Length of Stay ($>$7 days; 14.1\%), (E3)~ICU Mortality (7.4\%), and (E4)~Post-Observation Mortality (11.2\%). All MIMIC-IV results report mean $\pm$ std across 5~random seeds unless otherwise noted.
 
\paragraph{eICU.}
To test cross-center generalization, we evaluate on the eICU Collaborative Research Database~\cite{pollard2018eicu}, spanning 208~hospitals across the United States. We apply the same pipeline with no architectural modifications.
 
\paragraph{LEMMA-RCA.}
To test domain transfer beyond healthcare, we evaluate on LEMMA-RCA~\cite{zheng2024lemma}, a microservice fault detection benchmark (prevalence 22\%). Time-series evidence units correspond to service-level metrics and text units to log message chunks. The ToE pipeline is applied without modification.

\paragraph{Reproducibility and Hyperparameters.}
For the ToE beam search, we set beam width $B = 8$, maximum search depth $S_{\max} = 10$, and restrict candidates to the top $N_{\text{ts}} = 24$ time-series windows and $N_{\text{note}} = 20$ note chunks per instance. The search objective weights are $\lambda = 1.0$ (stability) and $\mu = 0.05$ (sparsity cost), with stopping thresholds $\tau_{\text{conf}} = 0.9$ and $\tau_{\text{suff}} = 0.9$. All experiments use a batch size of 32 on a single NVIDIA A100 GPU.

\subsection{Baseline Comparisons}
\label{sec:baselines}
We compare ToE against six baselines: \textbf{Top-$K$ Ranking}, which selects the $k$ units with the highest individual selector scores; \textbf{Saliency (Gradient)}, which ranks units by Input~$\times$~Gradient magnitude; \textbf{LIME}~\cite{ribeiro2016should} and \textbf{SHAP}~\cite{lundberg2017unified}, which select the top-$k$ units by local surrogate coefficients and Shapley-value attributions, respectively; a \textbf{Hard Concept Bottleneck Model (CBM)}~\cite{koh2020concept} with 24~binary clinical concepts grounded in established scoring systems; and \textbf{Random} selection as a lower bound.

\paragraph{Faithfulness--Sparsity Frontier.} Figure~\ref{fig:pareto_frontier} and Table~\ref{tab:lime_shap} compare all methods across evidence budgets. ToE matches the full model's predictive power (AUROC $\approx 0.800$) with as few as $k{=}5$ units (Fig.~\ref{fig:pareto_frontier}a) while maintaining the lowest fidelity error and ECE at every sparsity level (Fig.~\ref{fig:pareto_frontier}b). At $k{=}1$, ToE reduces fidelity Mean Absolute Error (MAE) by 39\% relative to Top-$k$ Ranking (0.158 to 0.096) and by 58\% relative to LIME, and outperforms LIME by 22~AUROC points. SHAP is the strongest attribution baseline, matching ToE's AUROC and MAE at $k \geq 5$, but exhibits 22\% higher fidelity MAE at $k=1$ ($0.123$ vs. $0.096$) and higher ECE at every budget, indicating weaker calibration alignment. ToE, by explicitly optimizing for probability stability (Eq.~\ref{eq:stability}), captures the model's actual confidence rather than just label correlations. The gap between methods narrows at higher budgets as the evidence space saturates. Multi-seed ECE results across all four MIMIC-IV tasks confirm that ToE achieves comparable or lower calibration error than the full model (Appendix~\ref{app:lime_shap_full}; Figure~\ref{app:ece}).

\begin{table}[t]
\centering
\small
\caption{Comparison with LIME and SHAP (E1: Hospital Mortality, 5~seeds). ToE achieves the best fidelity--sufficiency tradeoff at every budget $k$. Full results across all $k$ in Appendix~\ref{app:lime_shap_full}, Figure~\ref{app:ece}.}
\label{tab:lime_shap}
\resizebox{\linewidth}{!}{
\begin{tabular}{clccc}
\toprule
$k$ & Method & AUROC & Fidelity MAE ($\downarrow$) & ECE ($\downarrow$) \\
\midrule
\multirow{3}{*}{1}
  & LIME & 0.564 $\pm$ 0.006 & 0.229 $\pm$ 0.022 & 0.406 $\pm$ 0.011 \\
  & SHAP & 0.764 $\pm$ 0.009 & 0.123 $\pm$ 0.006 & 0.320 $\pm$ 0.010 \\
  & \textbf{ToE}  & 0.783 $\pm$ 0.013 & 0.096 $\pm$ 0.005 & 0.297 $\pm$ 0.019 \\
\midrule
\multirow{3}{*}{5}
  & LIME & 0.695 $\pm$ 0.010 & 0.171 $\pm$ 0.016 & 0.332 $\pm$ 0.018 \\
  & SHAP & 0.801 $\pm$ 0.014 & 0.039 $\pm$ 0.002 & 0.302 $\pm$ 0.025 \\
  & \textbf{ToE}  & 0.800 $\pm$ 0.017 & 0.040 $\pm$ 0.003 & 0.280 $\pm$ 0.023 \\
\midrule
\multirow{3}{*}{10}
  & LIME & 0.743 $\pm$ 0.011 & 0.126 $\pm$ 0.011 & 0.308 $\pm$ 0.020 \\
  & SHAP & 0.802 $\pm$ 0.017 & 0.024 $\pm$ 0.002 & 0.299 $\pm$ 0.029 \\
  & \textbf{ToE}  & 0.803 $\pm$ 0.017 & {0.035} $\pm$ 0.002 & 0.283 $\pm$ 0.025 \\
\bottomrule
\end{tabular}}
\end{table}

\paragraph{Comparison with Concept Bottleneck Models.}
The Hard CBM with 24~clinical concepts achieves AUROC 0.775 (AUPRC 0.349). ToE exceeds CBM on AUROC with a single evidence unit ($k = 1:$ $0.783$) and widens the gap at $ k = 5$ ($0.800$), while requiring no predefined concept annotations — though CBM retains an AUPRC edge from its curated concepts. This highlights a key advantage: CBMs require domain experts to define a fixed concept vocabulary before training, whereas ToE discovers relevant evidence units from learned representations at inference time.

\paragraph{Comparison with LLMs.}
We compare against 10 open-source (M)LLMs (1B–70B parameters), including medical fine-tunes and vision-language models, evaluated on E1 via zero-shot prompting with the full test set. Table~\ref{tab:llm} reports a representative subset; full results are in  Appendix~\ref{app:llm_full}; Figure~\ref{fig:llm}. Even the strongest model, Med42-v2-70B (AUROC of 0.745), underperforms ToE at $k = 5$ (AUROC of 0.800) despite having $640\times$ more parameters. Vision-language models (Gemma-2-12B (V), MedGemma-27B (V)~\cite{team2024gemma}) underperform their text-only counterparts on this task, suggesting that current Multimodal Large Language Models (MLLMs) struggle to extract discriminative signals from raw clinical images for structured prediction.  

\begin{table}[t]
\centering
\small
\caption{LLM/MLLM comparison (E1: Hospital Mortality). ToE with 109M parameters outperforms all models up to 70B. Full 10-model results in Appendix~\ref{app:llm_full}; Figure~\ref{fig:llm}.}
\label{tab:llm}
\resizebox{\linewidth}{!}{
\begin{tabular}{llccc}
\toprule
Model & Type & Params & AUROC & AUPRC \\
\midrule
Llama{-}3.2{-}1B & text & 1.2B & $0.532 \pm 0.009$ & $0.135 \pm 0.005$ \\
Llama{-}3.1{-}8B & text & 8.0B & $0.691 \pm 0.008$ & $0.206 \pm 0.008$ \\
Med42{-}v2{-}70B & text & 70B & $0.745 \pm 0.009$ & $0.293 \pm 0.014$ \\
MedGemma{-}27B (V) & vision & 27B & $0.630 \pm 0.009$ & $0.190 \pm 0.008$ \\
\midrule
\textbf{ToE} $k{=}5$ & multi & 109M & 0.800 $\pm$ 0.017 & 0.310 $\pm$ 0.067 \\
\bottomrule
\end{tabular}}
\end{table}

\subsection{Cross-Task and Cross-Dataset Generalization}
\label{sec:generalization}
 
A primary concern is whether ToE generalizes beyond a single task and dataset. Table~\ref{tab:generalization} reports results across all six evaluation settings.
 
\begin{table}[t]
\centering
\small
\caption{Cross-task and cross-dataset evaluation. ToE retains $\geq$98\% of full-model AUROC at $k{=}5$ across all settings. MIMIC-IV results: mean $\pm$ std over 5 seeds; eICU and LEMMA-RCA results are from a single seed.}
\label{tab:generalization}
\resizebox{\linewidth}{!}{
\begin{tabular}{llcccc}
\toprule
 & Dataset & Task & Full AUROC & ToE $k{=}5$ & Fid.\ MAE \\
\midrule
E1 & MIMIC-IV & Hosp.\ mort. & 0.806 $\pm$ 0.015 & 0.800 $\pm$ .017 & 0.040 $\pm$ 0.003 \\
E2 & MIMIC-IV & Long LOS     & 0.747 $\pm$ 0.041 & 0.740 $\pm$ .046 & 0.031 $\pm$ 0.002 \\
E3 & MIMIC-IV & ICU mort.    & 0.816 $\pm$ 0.009 & 0.808 $\pm$ .011 & 0.042 $\pm$ 0.004 \\
E4 & MIMIC-IV & Post-obs.    & 0.794 $\pm$ 0.021 & 0.784 $\pm$ .023 & 0.041 $\pm$ 0.001 \\
\midrule
   & eICU     & ICU mort.    & 0.822            & 0.808            & 0.124 \\
   & LEMMA    & Fault det.   & 0.741            & 0.730            & 0.106 \\
\bottomrule
\end{tabular}}
\end{table}
 
Three observations merit emphasis. First, ToE retains 98.5--99.3\% of full-model AUROC at $k = 5$ across all four MIMIC-IV tasks, with fidelity MAE consistently in the narrow range 0.031--0.042, despite substantial differences in clinical semantics and class balance (7.4--14.1\%). Second, eICU replicates the core finding on an independent multi-center dataset spanning 208~hospitals with different EHR systems and documentation practices. Third, LEMMA-RCA demonstrates that the same pipeline generalizes beyond healthcare entirely, with no architectural modifications.

\begin{table}[h]
\centering
\caption{\textbf{Modality Ablation Results ($k=5$).} Notes-Only fails to ground predictions, while the Multimodal (Both) approach maintains high predictive power and stability comparable to the strong TS-Only baseline.}
\label{tab:modality_ablation}
\resizebox{\linewidth}{!}{
\begin{tabular}{lccc}
\toprule 
\textbf{Modality} & \textbf{AUROC} & \textbf{AUPRC} & \textbf{Fidelity MAE}  \\
\midrule
TS Only    & $0.7876 \pm 0.0075$ & $0.3912 \pm 0.0151$ & $0.0445 \pm 0.0730$\\
Notes Only & $0.5590 \pm 0.0077$ & $0.1338 \pm 0.0047$ & $0.3432 \pm 0.2915$ \\
Both       & $0.8001 \pm 0.0165$ & $0.3096 \pm 0.0672$ & $0.0403 \pm 0.0027$ \\ 
\bottomrule
\end{tabular}}
\end{table}

\subsection{Ablation Studies}
To validate the components of the ToE framework, we analyzed the contribution of different modalities and the scoring objective.

\paragraph{Modality Necessity.}
Table~\ref{tab:modality_ablation} examines modality contributions at $k{=}5$. The Notes-Only baseline fails (AUROC $\approx 0.56$, MAE $> 0.3$), confirming that radiology text alone is insufficient for grounding without physiological context. The multimodal approach matches the predictive power of the time-series backbone (AUROC $\approx 0.80$) while maintaining low fidelity error (MAE $\approx 0.04$), validating ToE's design of using robust vitals to anchor the search while selectively retrieving text for semantic explanation.

\paragraph{Search \& Objective Analysis.}
% Table~\ref{tab:search_ablation} validates the "System-2" design: removing the stability objective ($\lambda=0$) more than doubles the fidelity error, proving that faithful explanations require matching the model's calibration, not just maximizing confidence. Furthermore, ToE performance confirms the necessity of combinatorial search over simple ranking.
Table~\ref{tab:search_ablation} validates the deliberative search design: removing the stability objective ($\lambda = 0$) more than doubles the fidelity error, confirming that faithful explanations require matching the model's calibration, not just maximizing confidence.

\begin{table}[h]
\centering
\caption{\textbf{Search Objective Ablation ($k=5$, MIMIC-IV Mortality, 5 seeds).} The full ToE objective achieves the lowest Fidelity MAE. Removing stability ($\lambda{=}0$) doubles the error. At fixed $k$, the sparsity cost $\mu$ has no effect (identical to Full). Top-$k$ Ranking without beam search doubles the MAE ($+100\%$).}
\label{tab:search_ablation}
\resizebox{\linewidth}{!}{
\begin{tabular}{lcccc}
\toprule
\textbf{Configuration} & \textbf{AUROC} & \textbf{AUPRC} & \textbf{Fidelity MAE} & \textbf{Comp.} \\
\midrule
Full ($\lambda{=}1.0$, $\mu{=}0.05$) & $0.8001 \pm 0.0165$ & $0.3096 \pm 0.0672$ & $0.0403 \pm 0.0027$ & $0.1112 \pm 0.0143$ \\
No Stability ($\lambda{=}0.0$, $\mu{=}0.05$) & $0.7738 \pm 0.0338$ & $0.3069 \pm 0.0622$ & $0.0800 \pm 0.0052$ & $0.1371 \pm 0.0153$ \\
No Sparsity ($\lambda{=}1.0$, $\mu{=}0.0$) & $0.7915 \pm 0.0385$ & $0.3191 \pm 0.0682$ & $0.0408 \pm 0.0015$ & $0.1025 \pm 0.0115$ \\
Top-$k$ Ranking (no search) & $0.7735 \pm 0.0339$ & $0.3066 \pm 0.0625$ & $0.0806 \pm 0.0054$ & $0.1357 \pm 0.0158$ \\
\bottomrule
\end{tabular}}
\end{table}

\paragraph{Search vs.\ Ranking.}
To isolate the contribution of combinatorial search from the scoring function, we compare beam search against Top-$k$ Ranking using identical selector scores (Appendix~\ref{app:search_rank}). The advantage is most pronounced under strict sparsity: at $k{=}1$, beam search achieves AUROC $0.783 \pm 0.013$ versus $0.768 \pm 0.036$ for ranking, with fidelity MAE reduced by $11\%$ ($0.096$ vs.\ $0.107$). At $k{=}5$, the gap widens to $+0.027$ AUROC and $50\%$ lower MAE ($0.040$ vs.\ $0.081$). The gap narrows at higher budgets as the evidence space saturates, confirming that combinatorial search matters most at sparse budgets where the model must identify the most decisive evidence units.

\subsection{Auditing Model Reliability via Search Behavior}
\label{sec:auditing}
ToE is faithful to the \emph{model's} logic, not to clinical ground truth; if the base predictor relies on spurious correlations or biases, ToE will faithfully surface them. We argue this is a feature rather than a limitation: ToE's search behavior provides a built-in diagnostic for prediction reliability.
 
\begin{table}[t]
\centering
\small
\caption{Search exhaustion rates for true positive (TP) vs.\ false positive (FP) predictions. When the model is wrong, the search exhausts its budget 4--26$\times$ more often.}
\label{tab:tp_fp}
\begin{tabular}{lcc}
\toprule
Metric & eICU & MIMIC-IV \\
\midrule
TP search exhaustion rate  & 0.3\%  & 7.2\%  \\
FP search exhaustion rate  & 7.3\%  & 30.2\% \\
FP / TP exhaustion ratio   & 25.6$\times$ & 4.2$\times$ \\
\bottomrule
\end{tabular}
\end{table}
 
Among positive predictions, we observe a systematic divergence between true positives and false positives (Table~\ref{tab:tp_fp}). When the model is correct, ToE finds supporting evidence almost immediately. When the model is wrong, the search struggles and exhausts its budget 4-26$\times$ more often. This asymmetry enables \emph{selective abstention}: flagging predictions where the search exhausted its budget catches 7.3\% of false positives on eICU while losing only 0.3\% of true positives, improving precision with negligible sensitivity loss. A spurious feature injection experiment further validates this signal: a model retrained with a deliberately spurious feature (80/20 correlated in training, 0\% in test) requires 4.5$\times$ more evidence to converge and halves its convergence rate (Appendix~\ref{app:spurious}).

\subsection{Efficiency Analysis}
A common concern with search-based methods is latency. Our timing analysis reveals that ToE adds only ${\sim}13$ms of overhead per patient compared to the full forward pass. This efficiency, achieved via caching BERT embeddings and lightweight GRU updates, makes ToE suitable for real-time clinical deployment.

\subsection{Qualitative Analysis}
\label{sec:qualitative}
To understand how ToE navigates the multimodal landscape, we visualize search traces for two representative patients in Table~\ref{tab:case_studies}.

\begin{table*}[ht]
\centering
\small
\caption{Qualitative comparison of ToE traces. \textbf{(Left)} ToE efficiently solves clear-cut cases using only vitals. \textbf{(Right)} ToE dynamically integrates clinical notes to surface interpretable clinical context, extracting specific medical concepts (e.g., ``Alveolar Edema'') to ground the prediction.}
\renewcommand{\arraystretch}{1.2}
\begin{tabular}{p{0.48\textwidth} | p{0.48\textwidth}}
\toprule \hline
\textbf{Case A: Efficient Triage} (Patient A) & \textbf{Case B: Multimodal Synergy} (Patient B) \\
\midrule
\textbf{Task:} Mortality Prediction & \textbf{Task:} Mortality Prediction \\
\textbf{Full Model Prob:} $0.0005$ (Low Risk) & \textbf{Full Model Prob:} $0.861$ (High Risk) \\
\textbf{Final Trace:} $k=1$ (Vitals Only) & \textbf{Final Trace:} $k=10$ (8 Vitals + 2 Notes) \\
\midrule
\textbf{Search Step 1:} Add Vitals \texttt{W5} \newline
$\hookrightarrow$ \textit{Evidence:} [Physiological Window 5] \newline
$\hookrightarrow$ \textit{Sufficiency:} $\mathbf{0.998}$ (Threshold Met) & 
\textbf{Search Steps 1--8:} Add Vitals \texttt{W23, W1, \ldots, W11} \newline
$\hookrightarrow$ \textit{Evidence:} [Multiple Physiological Windows] \newline
$\hookrightarrow$ \textit{Sufficiency:} $0.840$ (Plateau) \\
\midrule
\textbf{Outcome:} The search terminates immediately. The model determines that the vital signs alone are sufficient to justify the ``Low Risk'' prediction. No notes are processed. & 
\textbf{Search Step 9:} Add Note \texttt{N2} \newline
$\hookrightarrow$ \textit{Evidence:} \textbf{``\ldots Coalescent, bilateral, perihilar opacities reflect alveolar edema\ldots\ suggest volume overload.''} \newline
$\hookrightarrow$ \textit{Sufficiency:} $0.833$ (Stable) \\
\midrule
& \textbf{Search Step 10:} Add Note \texttt{N3} \newline
$\hookrightarrow$ \textit{Evidence:} [Additional Radiology Report] \newline
$\hookrightarrow$ \textit{Sufficiency:} $0.833$ (No Change) \\
\midrule
& \textbf{Outcome:} Vitals carry the primary predictive signal but plateau below the sufficiency threshold. The search retrieves the ``Volume Overload'' finding to provide clinically interpretable grounding for the high mortality risk. The slight sufficiency dip ($0.840 \to 0.833$) falls within noise; since $\tau_{suff} = 0.9$ is unmet, the search continues and surfaces cross-modal evidence. \\ \hline
\bottomrule
\end{tabular}
\label{tab:case_studies}
\end{table*}

\paragraph{Case A: Efficient Triage (Vitals-Only).}
For Patient A, the model identifies a clear physiological deterioration solely from time-series data. The search selects a single vital-sign window (\texttt{W5}) showing acute instability. This evidence alone yields a sufficiency score of $0.998$, triggering the stopping criterion immediately ($k=1$). By recognizing that the vitals are unambiguous, ToE avoids processing the clinical notes entirely, reducing computational cost.

\paragraph{Case B: Multimodal Resolution.}
In contrast, Patient B presents a more complex picture. The search initially retrieves multiple time-series windows (W23, W1, ..., W11), but the sufficiency score plateaus around $0.84$, indicating the physiological signals alone do not fully explain the model's high-risk prediction.  At this plateau, ToE expands to the clinical notes, retrieving a specific radiology report segment (\texttt{N2})  that documents \emph{"...bilateral perihilar opacities reflect alveolar edema... suggest volume overload."} While sufficiency remains stable ($0.833$), this textual evidence provides the causal context \emph{(Volume Overload)} that grounds the physiological signals in a clinically interpretable explanation — demonstrating ToE's ability to surface relevant cross-modal evidence even when the vitals alone carry the predictive signal.

\begin{table*}[t]
\centering
\small
\setlength{\tabcolsep}{5pt}
\caption{\textbf{ToE vs Zero-Shot LLM.} Comparison on a representative subset of 5 patients. While the LLM achieves perfect prediction accuracy by leveraging external medical knowledge, ToE remains faithful to the underlying model's logic. ToE selects significantly sparser evidence (Avg 6.2 vs. 9.0 windows) while maintaining reasonable overlap with the LLM's clinical reasoning.}
\label{tab:llm_comparison}
\resizebox{\linewidth}{!}{
\begin{tabular}{lcccccccl}
\toprule \hline
\textbf{Patient ID} & \textbf{Outcome} & \multicolumn{2}{c}{\textbf{Prediction}} & \multicolumn{2}{c}{\textbf{Evidence Size ($k$)}} & \textbf{Jaccard} & \textbf{Key Insight} \\
\cmidrule(lr){3-4} \cmidrule(lr){5-6}
& & \textbf{ToE} & \textbf{LLM} & \textbf{ToE} & \textbf{LLM} & \textbf{Overlap} & \\
\midrule
1 & Survived & \cmark & \cmark  & \textbf{1} & 5 & 20.0\% & ToE solved via single vital; LLM was cautious. \\
2 & Survived & \cmark  & \cmark  & \textbf{6} & 9 & 50.0\% & Strong agreement on deterioration intervals. \\
3 & Died & \cmark  & \cmark  & \textbf{8} & 11 & 46.2\% & Both flagged critical physiological decline. \\
4 & Survived & \cmark & \cmark  & \textbf{8} & 9 & 30.8\% & LLM prioritized stable periods differently. \\
5 & Died & \xmark & \cmark  & \textbf{8} & 11 & 35.7\% & \textbf{Audit Win:} ToE exposed model blindness to GCS. \\
\midrule
\textbf{Average} & - & 80\% Acc & 100\% Acc & \textbf{6.2} & 9.0 & 36.5\% & ToE is $\sim30\%$ more sparse than LLM. \\
\bottomrule
\bottomrule
\end{tabular}}
\end{table*}

\paragraph{Comparison with Zero-Shot LLM Evidence Selection.}
To contextualize ToE evidence selection against a strong "clinician-like" baseline, we compare ToE to a zero-shot LLM that selects hourly evidence windows from the same 24-hour observation period. We summarize results on ICU stays where both methods produce an explicit set of windows.\footnote{We use "evidence size" to denote the number of selected hourly windows. For ToE, this corresponds to the final evidence set returned by the beam search. For the LLM, this corresponds to the set of windows it explicitly marked as supporting evidence.} 
Across these cases, ToE achieves higher predictive accuracy than the LLM (0.655 ± 0.064 vs. 0.619 ± 0.070), while also selecting similarly small evidence sets on average (5.0 ± 0.0 vs. 4.9 ± 1.1 windows). To quantify agreement between the two evidence traces, we compute Jaccard similarity between the selected window sets. Agreement remains modest overall, with a mean Jaccard similarity of 0.125 ± 0.106 for time-series evidence, 0.310 ± 0.462 for clinical notes, and 0.112 ± 0.090 when combining all modalities.
Table~\ref{tab:llm_comparison} highlights five patients, and shows a consistent \emph{sparsity--context} tradeoff: ToE selects fewer evidence windows on average while maintaining non-trivial overlap with the LLM’s selections (mean Jaccard 0.365). Importantly, when the model is wrong, ToE’s trace remains valuable because it reveals \emph{which evidence the model actually relied on}, enabling targeted auditing.

A critical divergence occurred with Patient 5 (Died), where the LLM correctly predicted "High Risk" by identifying persistent neurological failure (Glasgow Coma Scale (GCS) 8-10). In contrast, ToE faithfully revealed that the underlying EB model predicted "Low Risk" because it prioritized stable respiratory signals (SpO2 $\sim$100\%) and ignored the GCS trajectory.  This failure case highlights the danger of using LLMs as explanations: the LLM "hallucinated" a correct reasoning path that the model did not actually use. ToE, by contrast, successfully exposed the model's blind spot regarding neurological status.
\section{Conclusion}
\label{sec:conclusion}
% We introduced ToE, an inference-time search framework for generating faithful multimodal rationales for prediction models. By formulating interpretability as a discrete optimization problem over evidence units and combining Evidence Bottlenecks with beam search, ToE produces auditable traces that explicitly identify which evidence units support a model's prediction. Our experiments on ICU mortality prediction in MIMIC-IV show that ToE preserves both the full model decision and its predicted probability under strict evidence budgets, and improves probability fidelity relative to greedy and saliency-based baselines when only a small number of evidence units may be selected. These results indicate that search-based rationale extraction can more accurately recover the model's own decision logic than methods based solely on evidence ranking. Future work will extend ToE to additional tasks and data modalities.
We introduced ToE, an inference-time search framework for generating faithful multimodal rationales. By formulating interpretability as a discrete optimization problem over evidence units and combining Evidence Bottlenecks with beam search, ToE produces auditable traces that identify which evidence units support a model's prediction. Across six tasks spanning three datasets (MIMIC-IV, eICU, LEMMA-RCA) and two domains, ToE retains at least 98\% of full-model AUROC with as few as five evidence units. Under sparse evidence budgets, ToE achieves lower fidelity error than LIME, SHAP, Top-$k$ Ranking, and greedy baselines, and outperforms CBMs without requiring predefined annotations. ToE also outperforms LLMs up to 70B parameters on clinical prediction tasks with a 109M-parameter model. Beyond explanation, ToE's search behavior provides a practical diagnostic for prediction reliability: search exhaustion rates are 4-26$\times$ higher for false positives than true positives, enabling selective abstention. These results indicate that search-based rationale extraction can more accurately recover a model's decision logic than methods based on evidence ranking or post-hoc attribution. ToE is currently validated on late-fusion architectures with separable evidence streams. Extending the framework to cross-attention and early-fusion models, for example, through attention-head decomposition or adapter layers, is an important direction for future work.

\section{Limitations}
\label{sec:limitations}
ToE produces evidence sets that are faithful to the underlying model's decision logic, not to clinical ground truth. If the base predictor relies on spurious correlations or biases, ToE will surface them rather than correct them, though, as shown in Section~\ref{sec:auditing}, this model-faithfulness itself serves as a diagnostic for unreliable predictions. More broadly, ToE is an interpretability wrapper: it cannot fix errors in the base model, and its coarse evidence units (hourly windows, report chunks) may omit finer-grained clinically relevant signals.

The framework is currently validated on late-fusion architectures with separable evidence streams; extending to cross-attention or early-fusion models requires additional design. Beam search finds near-optimal evidence sets under the scoring heuristic, not globally optimal ones, though exhaustive enumeration confirms gaps below 0.001 AUROC at small $k$ on MIMIC-IV (Appendix~\ref{app:optimality}). Finally, while ToE's $\sim$ 13ms overhead is practical for most settings, runtime may increase with longer note histories or larger beam widths.

\section{Ethical Considerations}
\label{sec:ethics}
\paragraph{Clinical Safety and Intended Use.}
This research presents a prototype for clinical decision support and is not intended for autonomous diagnosis or treatment planning. False negatives in mortality prediction could lead to reduced care, while false positives could cause alarm fatigue. ToE is designed explicitly to mitigate these risks by forcing the model to show its work, allowing clinicians to verify or reject the machine's rationale. We emphasize that the selected evidence is a mathematical construct reflecting the model's confidence, not a comprehensive summary of the patient's clinical state.

\paragraph{Data Privacy and Compliance.}
Our models were developed using the MIMIC-IV dataset, which contains de-identified electronic health records from Beth Israel Deaconess Medical Center. We adhered to the PhysioNet Credentialed Data Use Agreement, ensuring no attempt was made to re-identify patients. Any deployment of this technology in a live clinical setting would require strict adherence to local regulations (e.g., HIPAA in the US, GDPR in Europe) and rigorous external validation.

\paragraph{Bias and Fairness.}
Clinical datasets are known to harbor demographic and socioeconomic biases. A model trained on MIMIC-IV (collected in Boston, MA) may underperform or rely on different feature sets for underrepresented populations. A key advantage of ToE is its ability to \emph{audit} these biases; by inspecting the evidence trees, stakeholders can detect if the model relies on impermissible proxies (e.g., insurance status or language barriers) for its predictions. However, the search algorithm itself does not remove these biases, and deploying the model without fairness audits could perpetuate existing healthcare disparities.

\section*{Acknowledgments}
% This research was supported in part through research cyber-infrastructure resources and services, including the AI Makerspace of the College of Engineering, provided by the Partnership for an Advanced Computing Environment (PACE) at the Georgia Institute of Technology, Atlanta, Georgia, USA. 
This research was supported in part through research cyberinfrastructure resources and services, including the AI Makerspace of the College of Engineering (RRID:SCR\_028058), provided by the Partnership for an Advanced Computing Environment (PACE) at the Georgia Institute of Technology, Atlanta, Georgia, USA (RRID:SCR\_027619). We also gratefully acknowledge funding and fellowships that contributed to this work, including a Wallace H. Coulter Distinguished Faculty Fellowship, a Petit Institute Faculty Fellowship, and research funding from Amazon and Microsoft Research awarded to Professor May D. Wang.
\bibliography{custom}
\appendix
\renewcommand{\thefigure}{A\arabic{figure}}
\renewcommand{\thetable}{A\arabic{table}}
\setcounter{figure}{0}
\setcounter{table}{0}
\section{Appendix}
\label{sec:appendix}
\subsection{Detailed Performance Across Budgets}
Figure~\ref{app:ece} and Table~\ref{tab:comprehensiveness} detail the performance of ToE across varying evidence budgets ($k$). Notably, Sufficiency AUROC saturates at $k=5$, indicating that a handful of clinical events are often sufficient for robust diagnosis.

\begin{table}
\centering
\caption{ToE Comprehensiveness ($\uparrow$) across evidence budgets and four MIMIC-IV tasks (mean $\pm$ std, 5 seeds).}
\label{tab:comprehensiveness}
\resizebox{\linewidth}{!}{%
\begin{tabular}{ccccc}
\toprule
$k$ & \textbf{E1: Hospital Mort.} & \textbf{E2: Long LOS} & \textbf{E3: ICU Mort.} & \textbf{E4: Post-Obs Mort.} \\
\midrule
1 & $0.0413 \pm 0.0050$ & $0.0447 \pm 0.0107$ & $0.0326 \pm 0.0050$ & $0.0441 \pm 0.0037$ \\
3 & $0.0885 \pm 0.0104$ & $0.0939 \pm 0.0094$ & $0.0749 \pm 0.0092$ & $0.0908 \pm 0.0051$ \\
5 & $0.1112 \pm 0.0143$ & $0.1199 \pm 0.0104$ & $0.0961 \pm 0.0131$ & $0.1139 \pm 0.0067$ \\
8 & $0.1293 \pm 0.0177$ & $0.1431 \pm 0.0117$ & $0.1138 \pm 0.0159$ & $0.1334 \pm 0.0089$ \\
12 & $0.1417 \pm 0.0199$ & $0.1582 \pm 0.0125$ & $0.1270 \pm 0.0180$ & $0.1463 \pm 0.0108$ \\
16 & $0.1489 \pm 0.0216$ & $0.1644 \pm 0.0123$ & $0.1342 \pm 0.0190$ & $0.1541 \pm 0.0121$ \\
20 & $0.1549 \pm 0.0233$ & $0.1681 \pm 0.0119$ & $0.1390 \pm 0.0201$ & $0.1596 \pm 0.0137$ \\
24 & $0.1612 \pm 0.0186$ & $0.1616 \pm 0.0097$ & $0.1307 \pm 0.0403$ & $0.1539 \pm 0.0287$ \\
\bottomrule
\end{tabular}%
}
\end{table}
\subsection{Calibration Analysis}
\label{app:calibration}
\begin{figure}
    \centering
    \includegraphics[width=\linewidth]{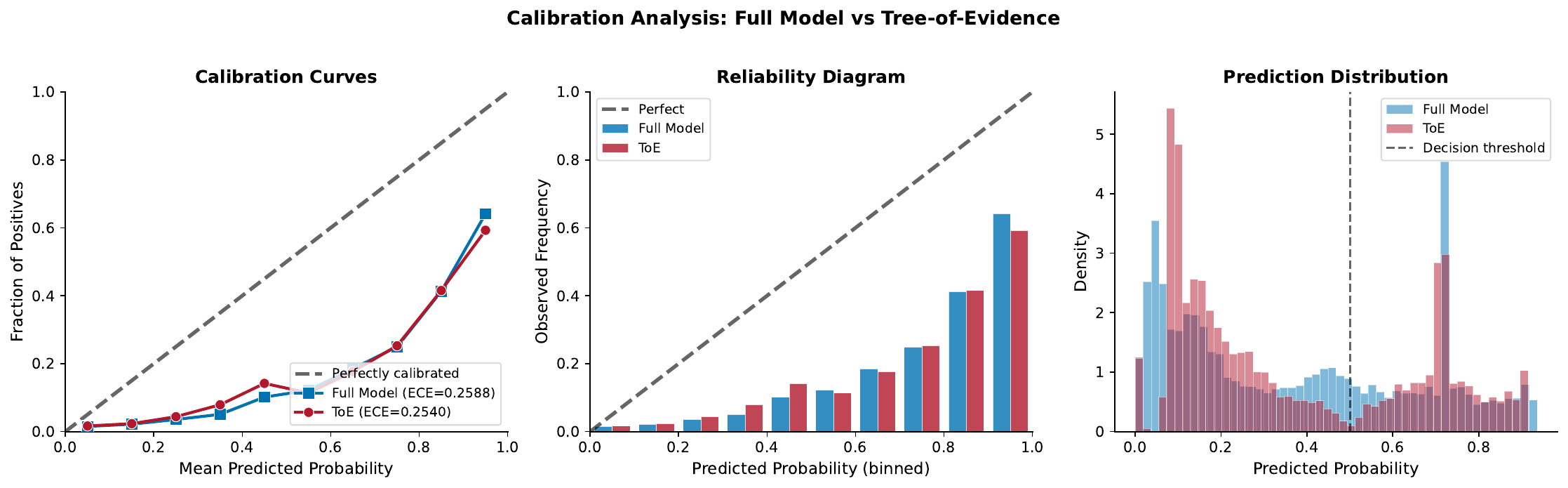}
    \caption{\textbf{Calibration Analysis.} Full Model (Blue) vs.\ ToE at $k{=}5$ (Red). ToE preserves calibration (ECE 0.254 vs.\ 0.259) while using sparse evidence.}
    \label{fig:calibration}
\end{figure}

We evaluated calibration using Expected Calibration Error (ECE). As shown in Figure~\ref{fig:calibration}, ToE achieves comparable calibration to the full model (ECE 0.254 vs.\ 0.259), with both models exhibiting similar reliability curves across probability bins. The prediction distributions confirm that ToE preserves the full model's confidence profile while operating on only $k{=}5$ evidence units.

\subsection{Evidence Size Distribution}
\label{app:evidence_dist}
\begin{figure}[ht]
    \centering
    \includegraphics[width=\linewidth]{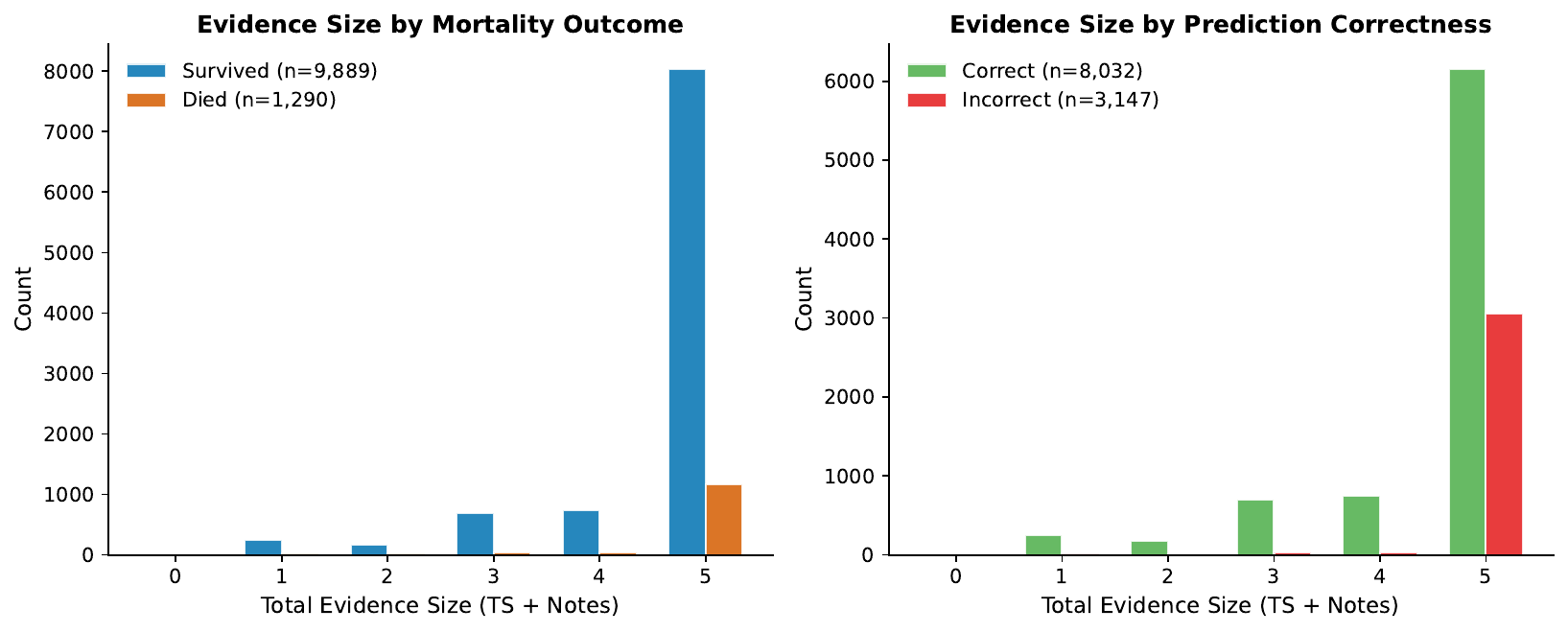}
    \caption{\textbf{Evidence Size Distribution ($k{=}5$, MIMIC-IV Mortality).} \textbf{Left:} By mortality outcome, both groups concentrate at the budget cap, with survivors showing slightly more early stopping. \textbf{Right:} By prediction correctness, correct predictions exhibit greater evidence efficiency (more mass at $k < 5$), while incorrect predictions exhaust the full budget in 97\% of cases, indicating the search struggles to find supporting evidence when the model is wrong.}
    \label{fig:evidence_dist}
\end{figure}

Figure~\ref{fig:evidence_dist} illustrates the distribution of selected evidence sizes at budget $k{=}5$, stratified by patient outcome and prediction correctness. When the model is correct (n=8{,}032), ToE frequently finds sufficient evidence before exhausting the budget, with notable mass at $k{=}1$--$4$. When the model is incorrect (n=3{,}147), the search almost universally consumes the full budget ($k{=}5$), reflecting the absence of a coherent evidence subset that supports the (wrong) prediction. This asymmetry makes evidence utilization a diagnostic signal for prediction reliability.

\subsection{Tree of Evidence (ToE) Algorithm}
\label{app:toe_algo}
We provide below an algorithm for the ToE (Algorithm~\ref{alg:toe}).
\begin{algorithm}[ht]
\caption{Tree-of-Evidence (ToE) Inference Search}
\label{alg:toe}
\begin{algorithmic}[1]
\Require Trained models; Candidates $\mathcal{W}, \mathcal{N}$; Beam width $B$
\State Compute target $p_{\text{full}}$ and $\hat{y}_{\text{full}}$ using all data
\State Cache note chunk embeddings $\{e_j\}$
\State $Beam \gets [(\mathbf{0}^{\text{ts}}, \mathbf{0}^{\text{note}})]$
\For{$step = 1$ to $S_{\max}$}
    \State $Candidates \gets \emptyset$
    \For{state $(m^{\text{ts}}, m^{\text{note}})$ in $Beam$}
        \State Expand by adding one unused $w \in \mathcal{W}$ or $n \in \mathcal{N}$
        \State Compute $\text{score}(\cdot)$ via Eq.~(\ref{eq:score})
        \State $Candidates \gets Candidates \cup \{\text{NewState}\}$
    \EndFor
    \State $Beam \gets \text{Top-}B(Candidates)$
    \If{$Beam[0]$ meets thresholds $\tau_{\text{conf}}, \tau_{\text{suff}}$}
        \State \Return $Beam[0]$ \Comment{Minimal Sufficient Set}
    \EndIf
\EndFor
\State \Return Best state in $Beam$
\end{algorithmic}
\end{algorithm}

\section{Full LIME/SHAP Comparison}
\label{app:lime_shap_full}
% ============================================================
 
\begin{figure*}[ht]
    \centering
    \includegraphics[width=\linewidth]{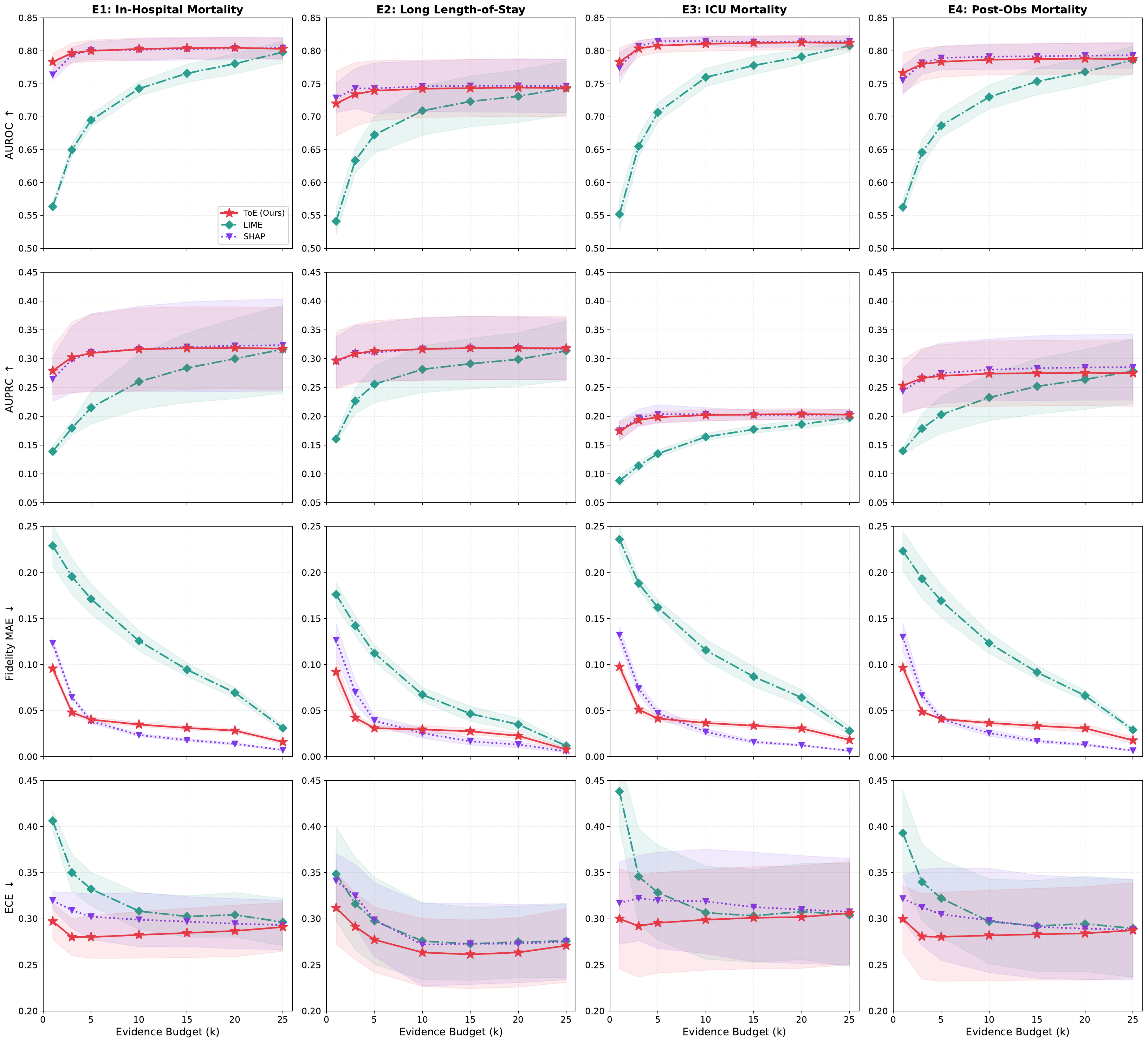}
    \caption{\textbf{Complete comparison of ToE, LIME, and SHAP across four MIMIC-IV tasks} (E1--E4, 5 seeds, $k \in \{1,3,5,10,15,20,25\}$). Rows: AUROC, AUPRC, Fidelity MAE, ECE. Columns: E1 (In-Hospital Mortality), E2 (Long LOS), E3 (ICU Mortality), E4 (Post-Obs Mortality). Shaded regions denote $\pm 1$ std. ToE consistently achieves the best ECE across all tasks and competitive AUROC at sparse budgets ($k \leq 5$), while SHAP converges at higher $k$ with lower MAE.}
    \label{app:ece}
\end{figure*}

Figure~\ref{app:ece} extends the main-paper comparison (Table~\ref{tab:lime_shap}) to all evidence budgets $k \in \{1, 3, 5, 10, 15, 20, 25\}$ with AUROC, AUPRC, Fidelity MAE, and ECE.

\section{Full LLM and CBM Comparison}
\label{app:llm_full}
% ============================================================
 \begin{figure}
    \centering
    \includegraphics[width=\linewidth]{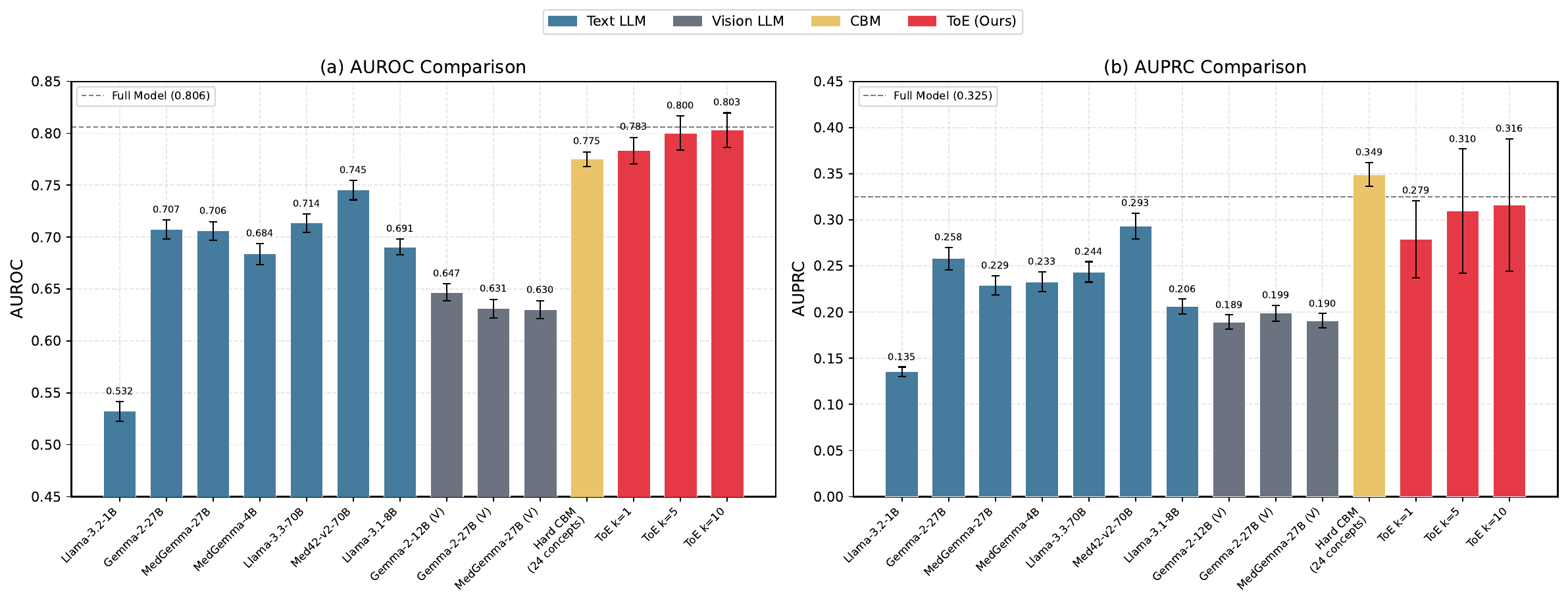}
    \caption{\textbf{LLM/MLLM, CBM, and ToE Comparison on E1 (In-Hospital Mortality, MIMIC-IV).} (a)~AUROC and (b)~AUPRC for 7 text-only LLMs (1B--70B), 3 vision LLMs (12--27B), Hard CBM (24 clinical concepts), and ToE at $k{=}1,5,10$ (109M parameters). Error bars denote $\pm 1$ std (bootstrap for LLMs/CBM, 5 seeds for ToE). Dashed line indicates the full model. ToE $k{=}5$ outperforms the best 70B LLM (Med42) by $+0.048$ AUROC with $640{\times}$ fewer parameters, and adding vision to LLMs degrades performance.}
    \label{fig:llm}
\end{figure}
Figure~\ref{fig:llm} reports the complete comparison against 10~open-source LLMs, CBM, and multimodal LLMs on E1 (Hospital Mortality), evaluated via zero-shot prompting on the full test set. All models are run locally via vLLM.

\section{Search vs.\ Ranking: Full Comparison}
\label{app:search_rank}
Table~\ref{tab:search_vs_ranking} provides the full comparison between ToE beam search and Top-$k$ Ranking across all evidence budgets (E1: Hospital Mortality).
\begin{table}[ht]
\centering
\caption{Search vs Ranking Comparison Across Evidence Budgets (MIMIC-IV Mortality, 5 seeds).}
\label{tab:search_vs_ranking}
\resizebox{\linewidth}{!}{
\begin{tabular}{ccccc}
\toprule
& \multicolumn{2}{c}{\textbf{Beam Search (ToE)}} & \multicolumn{2}{c}{\textbf{Top-$k$ Ranking}} \\
\cmidrule(lr){2-3} \cmidrule(lr){4-5}
$k$ & AUROC & Fidelity MAE & AUROC & Fidelity MAE \\
\midrule
1 & $0.7833 \pm 0.0128$ & $0.0958 \pm 0.0053$ & $0.7676 \pm 0.0362$ & $0.1073 \pm 0.0071$ \\
3 & $0.7967 \pm 0.0152$ & $0.0481 \pm 0.0023$ & $0.7739 \pm 0.0343$ & $0.0798 \pm 0.0054$ \\
5 & $0.8001 \pm 0.0165$ & $0.0403 \pm 0.0027$ & $0.7735 \pm 0.0339$ & $0.0806 \pm 0.0054$ \\
8 & $0.8028 \pm 0.0162$ & $0.0367 \pm 0.0023$ & $0.7722 \pm 0.0325$ & $0.0840 \pm 0.0056$ \\
12 & $0.8039 \pm 0.0167$ & $0.0333 \pm 0.0022$ & $0.7713 \pm 0.0314$ & $0.0859 \pm 0.0060$ \\
16 & $0.8046 \pm 0.0160$ & $0.0307 \pm 0.0022$ & $0.7723 \pm 0.0311$ & $0.0848 \pm 0.0061$ \\
20 & $0.8048 \pm 0.0156$ & $0.0284 \pm 0.0019$ & $0.7766 \pm 0.0319$ & $0.0783 \pm 0.0059$ \\
24 & $0.8043 \pm 0.0159$ & $0.0254 \pm 0.0016$ & $0.7837 \pm 0.0346$ & $0.0459 \pm 0.0047$ \\
\bottomrule
\end{tabular}}
\end{table}

% ============================================================
\section{STE Temperature Sensitivity}
\label{app:ste_temperature}
% ============================================================
 
Table~\ref{tab:app_ste} reports the effect of STE temperature $\tau_{\mathrm{STE}}$ on selector performance, with full retraining per temperature (eICU, ICU Mortality).
 
\begin{table}[ht]
\centering
\small
\caption{STE temperature sensitivity (eICU). Performance varies $<$1\% across a 50$\times$ range.}
\label{tab:app_ste}
\resizebox{\linewidth}{!}{
\begin{tabular}{cccc}
\toprule
$\tau_{\mathrm{STE}}$ & Suff.\ AUROC ($k{=}6$) & AUPRC ($k{=}6$) & Suff.\ AUROC ($k{=}1$) \\
\midrule
0.1 & 0.792 & 0.316 & 0.741 \\
0.5 & 0.797 & 0.321 & 0.740 \\
1.0 & 0.799 & 0.325 & 0.753 \\
2.0 & 0.799 & 0.319 & 0.745 \\
5.0 & 0.790 & 0.313 & 0.748 \\
\bottomrule
\end{tabular}}
\end{table}

% ============================================================
\section{Probability-Space vs.\ Logit-Space Stability}
\label{app:stability_space}
% ============================================================
 
Table~\ref{tab:app_stability} compares probability-space and logit-space definitions of the stability term $S(\mathbf{m})$ across evidence budgets (E1: Hospital Mortality, MIMIC-IV).
 
\begin{table}[ht]
\centering
\small
\caption{Probability-space vs.\ logit-space stability. Probability-space achieves consistently lower fidelity MAE.}
\label{tab:app_stability}
\begin{tabular}{clccc}
\toprule
$k$ & Space & AUROC & Fid.\ MAE & Comp. \\
\midrule
\multirow{2}{*}{1}
  & Probability & 0.755 & 0.090 & 0.029 \\
  & Logit       & 0.749 & 0.100 & 0.040 \\
\midrule
\multirow{2}{*}{5}
  & Probability & 0.773 & 0.030 & 0.097 \\
  & Logit       & 0.768 & 0.054 & 0.131 \\
\midrule
\multirow{2}{*}{12}
  & Probability & 0.773 & 0.014 & 0.130 \\
  & Logit       & 0.769 & 0.051 & 0.189 \\
\bottomrule
\end{tabular}
\end{table}
 
Probability-space stability yields 44\% lower fidelity MAE at $k{=}5$. Notably, logit-space MAE plateaus at $\sim$0.05 regardless of budget, whereas probability-space MAE continues decreasing with more evidence. This pattern is confirmed on eICU.

% ============================================================
\section{Optimality Gap Analysis}
\label{app:optimality}
% ============================================================
 
To assess how close beam search comes to the global optimum, we compare ToE against exhaustive enumeration at small $k$ where enumeration is tractable (Table~\ref{tab:app_optimality}).
 
\begin{table}
\centering
\small
\caption{Optimality gap: ToE vs.\ exhaustive search.}
\label{tab:app_optimality}
\resizebox{\linewidth}{!}{
\begin{tabular}{llcccc}
\toprule
Dataset & $k$ & ToE AUROC & Exhaustive & Gap \\
\midrule
MIMIC-IV  & 1 & 0.7550 & 0.7543 & +0.0007 \\
MIMIC-IV  & 3 & 0.7706 & 0.7697 & +0.0009 \\
LEMMA-RCA & all & 0.7181 & 0.7252 & $-0.0071$ \\
\bottomrule
\end{tabular}}
\end{table}
 
At $k{=}1$ and $k{=}3$, ToE matches the global optimum (gap $< 0.001$ AUROC, within bootstrap confidence intervals). Exhaustive search becomes infeasible for $k \geq 5$ ($>$1M subsets per patient at $k{=}5$ on MIMIC-IV).

% ============================================================
\section{Spurious Feature Injection}
\label{app:spurious}
% ============================================================
To test whether ToE's search behavior can detect model reliance on spurious features, we retrain the model with a deliberately spurious binary feature that is 80\% correlated with mortality in the training set but has 0\% correlation in the test set.
 
The corrupted model requires 4.5$\times$ more evidence to converge ($p < 0.001$) and has half the convergence rate (46\% vs.\ 93\%). Within the corrupted model, the asymmetry between flag=1 and flag=0 patients (55\% vs.\ 37\% convergence) reveals the specific source of bias. This confirms that ToE's search difficulty is a reliable signal for detecting spurious model reasoning.
 
% ============================================================
\section{Hyperparameter Sensitivity}
\label{app:hyperparameters}
% ============================================================
 
Table~\ref{tab:app_hyperparams} summarizes sensitivity to key hyperparameters beyond the $\lambda/\mu$ ablation reported in the main paper (Table~\ref{tab:search_ablation}).
 
\begin{table}
\centering
\small
\caption{Hyperparameter sensitivity summary.}
\label{tab:app_hyperparams}
\resizebox{\linewidth}{!}{
\begin{tabular}{lccc}
\toprule
Hyperparameter & Range & Dataset(s) & Key Finding \\
\midrule
Stability space & Prob vs.\ Logit & MIMIC + eICU & Prob-space 44\% lower MAE \\
$\tau_{\text{suff}}$ & .70, .80, .90, .95 & eICU & Higher $\rightarrow$ better fidelity \\
$\lambda$ (stability) & 0, 1.0 & MIMIC & $\lambda{=}0$ doubles MAE \\
\bottomrule
\end{tabular}}
\end{table}
 
At fixed evidence budgets, stopping thresholds ($\tau_{\text{conf}}, \tau_{\text{suff}}$) have zero effect since they only control dynamic stopping. The method is robust to stability-space choice and threshold values but sensitive to $\lambda$, which is a core design parameter.

\end{document}